\documentclass[final]{cvpr}

\usepackage{times}
\usepackage{epsfig}
\usepackage{graphicx}
\usepackage{amsmath}
\usepackage{amssymb}
\usepackage{adjustbox}
\usepackage[sort,nocompress]{cite}
\usepackage{booktabs}

\usepackage[pagebackref=true,breaklinks=true,colorlinks,bookmarks=false]{hyperref}
\def\confYear{NTIRE 2021}

\begin{document}
    
%%%%%%%%% TITLE
\title{NTIRE 2021 Challenge on High Dynamic Range Imaging: \\ Dataset, Methods and Results}

\author{Eduardo P\'erez-Pellitero$^\dagger$ \and Sibi Catley-Chandar$^\dagger$ \and  Ale\v{s} Leonardis$^\dagger$ \and Radu Timofte$^\dagger$ \and
% teams ranked 1st (Track 1, Track 2)
% NOAHTCV
Xian Wang \and Yong Li \and Tao Wang \and Fenglong Song \and 
% MegHDR
Zhen  Liu \and Wenjie Lin \and Xinpeng Li \and Qing Rao \and Ting Jiang \and Mingyan Han \and Haoqiang Fan \and Jian Sun \and Shuaicheng Liu \and 
% teams ranked 2nd 
% XPixel
Xiangyu  Chen \and Yihao Liu \and Zhengwen Zhang \and Yu Qiao \and Chao Dong \and 
% SuperArtifacts
Evelyn Yi Lyn Chee \and Shanlan Shen \and Yubo Duan \and 
% teams ranked 3rd
% BOE-IOT-AIBD
Guannan Chen \and Mengdi Sun \and Yan Gao \and Lijie Zhang \and
% NOAHTCV -- already listed in position 1
% teams ranked 4th
% CET CVLAB
Akhil K A \and Jiji C V \and
% teans ranked 5th
% CVRG
S M A Sharif \and Rizwan Ali Naqvi \and Mithun Biswas \and Sungjun Kim \and
% ZJU231
Chenjie Xia \and Bowen Zhao \and Zhangyu Ye \and Xiwen Lu \and Yanpeng Cao \and Jiangxin Yang \and Yanlong Cao \and
% Incomplete submission
% Samsung Bangalore Research
Green Rosh K S \and Sachin Deepak Lomte \and  Nikhil Krishnan \and B H Pawan Prasad
}
\maketitle

%%%%%%%%% ABSTRACT
\begin{abstract}
This paper reviews the first challenge on high-dynamic range (HDR) imaging that was part of the New Trends in Image Restoration and Enhancement (NTIRE) workshop, held in conjunction with CVPR 2021. This manuscript focuses on the newly introduced dataset, the proposed methods and their results. The challenge aims at estimating a HDR image from one or multiple respective low-dynamic range (LDR) observations, which might suffer from under- or over-exposed regions and different sources of noise. The challenge is composed by two tracks: In Track 1 only a single LDR image is provided as input, whereas in Track 2 three differently-exposed LDR images with inter-frame motion are available. In both tracks, the ultimate goal is to achieve the best objective HDR reconstruction in terms of PSNR with respect to a ground-truth image, evaluated both directly and with a canonical tonemapping operation.

\end{abstract}
{\let\thefootnote\relax\footnotetext{%
\hspace{-5mm}$\dagger$ Eduardo P\'erez-Pellitero (\texttt{e.perez.pellitero@huawei.com}), Sibi Catley-Chandar, Ale\v{s} Leonardis (Huawei Noah's Ark Laboratory) and Radu Timofte (ETH Zurich) are the NTIRE 2021 challenge organizers, while the other authors participated in the challenge. Appendix~\ref{ap:teams-and-affiliations} contains the authors’ teams and affiliations.\\
 \url{https://data.vision.ee.ethz.ch/cvl/ntire21/}}}
%%%%%%%%% BODY TEXT
\section{Introduction}
Current consumer-grade cameras struggle to capture scenes with varying illumination with a single exposure shot due to the inherent limitations of the imaging sensor, which suffers from saturation in high-irradiance regions and from uncertainty in the readings for low-light regions.

In recent years, advances in computational photography have enabled single-sensor cameras to acquire images with an extended dynamic range without the need of expensive, bulky and arguably more inconvenient multi-camera rigs, \eg~\cite{Tocci11, McGuire07, Froehlich14}. Generally, those algorithms exploit multiple LDR frames captured with different exposure values (EV) that are then fused into a single HDR image~\cite{Debevec97,Mertens07}, with some of those methods including frame alignment~\cite{kalantari17, sen12, wu18} or pixel rejection strategies~\cite{yan19}.

Convolutional Neural Networks (CNNs) have greatly advanced the state-of-the-art for HDR reconstruction, especially for complex dynamic scenes~\cite{kalantari17, wu18, prabhakar20, yan19}. Additionally, CNNs have opened a new path into single-image HDR imaging thanks to their ability to learn  complex and entangled vision tasks seamlessly, \eg denoising, camera response function estimation, image in-painting, high-frequency and detail hallucination~\cite{liu20}. Despite the ill-posed nature of the single-image HDR reconstruction, most current methods obtain plausible results that, if not as accurate as those reconstructed from multiframe LDR images, can be a good alternative when multiple frames are not available or can not be captured due to time constrains.

The NTIRE 2021 HDR Challenge aims at stimulating research for computational HDR imaging, and better understand the state-of-the-art landscape for both single and multiple frame HDR processing. It is part of a wide spectrum of associated challenges with the NTIRE 2021 workshop: non-homogeneous dehazing~\cite{ancuti2021ntire}, defocus deblurring using dual-pixel~\cite{abuolaim2021ntire}, depth guided image relighting~\cite{elhelou2021ntire}, image deblurring~\cite{nah2021ntire}, multi-modal aerial view imagery classification~\cite{liu2021ntire}, learning the super-resolution space~\cite{lugmayr2021ntire}, quality enhancement of heavily compressed videos~\cite{yang2021ntire}, video super-resolution~\cite{son2021ntire}, perceptual image quality assessment~\cite{gu2021ntire}, burst super-resolution~\cite{bhat2021ntire}, high dynamic range~\cite{perez2021ntire}. 

%-------------------------------------------------------------------------
\section{Challenge}
\begin{figure*}
\centering
\includegraphics[width=1.0\textwidth]{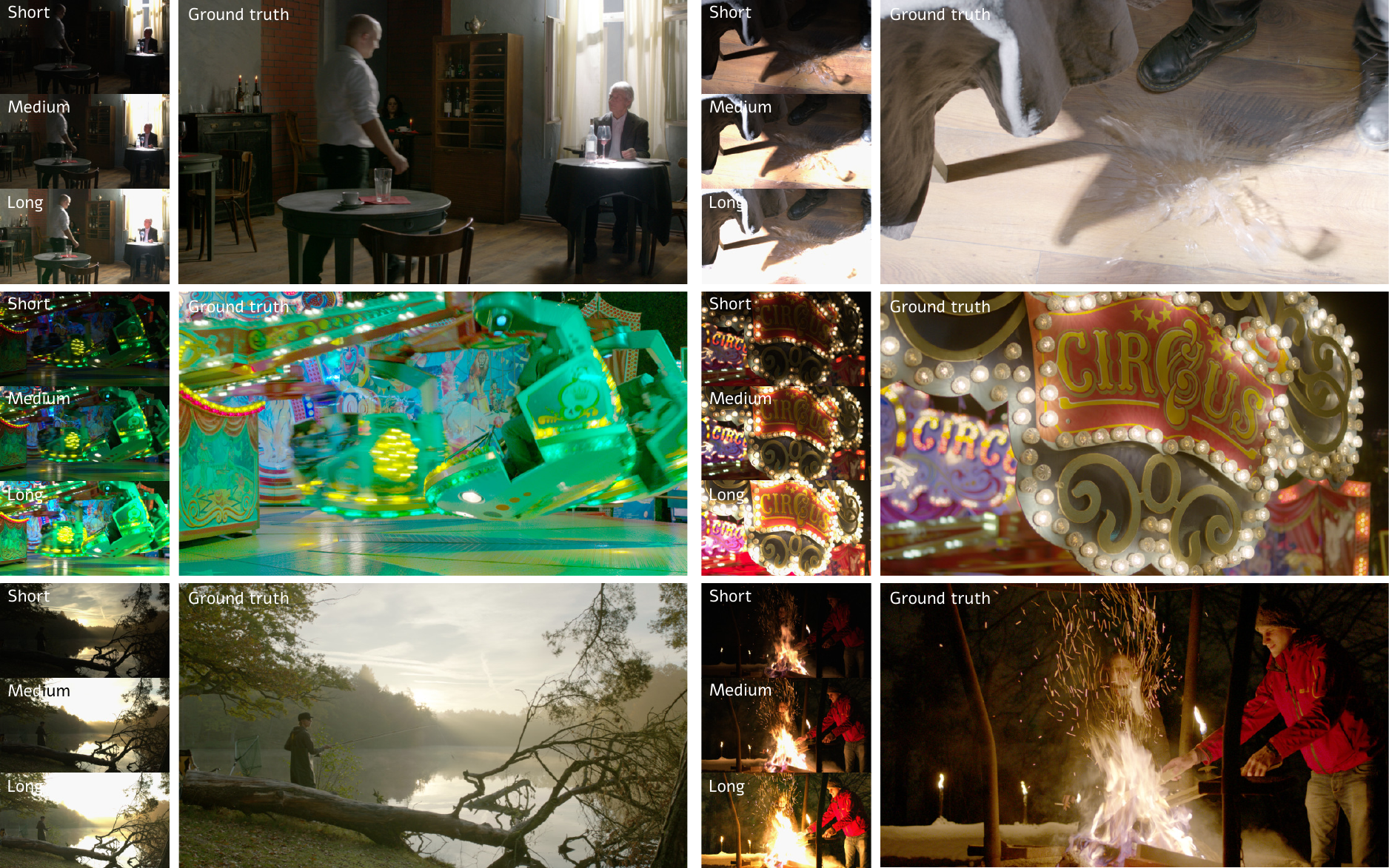}
\caption{Visualizations of samples included in the newly curated training, validation and testing datasets. Each training example is composed by three input LDR images (short, medium and long) and a related ground-truth image aligned with the medium input LDR. \textit{Note that validation and testing ground-true images were not made available to participants.}}
\label{fig:dataset-vis} 
\end{figure*}

The NTIRE 2021 HDR Challenge is the first edition that addresses the HDR image enhancement task. This challenge aims to gauge and advance the state-of-the-art on HDR imaging. It is focused specially in challenging scenarios for HDR image reconstruction, \ie wide range of scene illumination, accompanied by complex motions in terms of camera, scene and light sources. In this section we present details about the new dataset used for the challenge, as well as how the challenge tracks are designed.

\subsection{Dataset}

Both training and evaluation of HDR imaging algorithms require high quality annotated datasets. Specially for deep learning methods, the number of training examples and their diversity in terms of \eg scene and camera motion, exposure values, textures, semantic content, is of crucial importance for the model performance and generalization capabilities. Creating a high quality HDR dataset with such features still poses several challenges. Current HDR datasets are generally captured using static image bracketing, with some efforts towards controlling the scene motion so that \textit{stop-motion} dynamic scenes can be assembled: In the work of Kalantari \etal~\cite{kalantari17} a subject is asked to stay still in order to capture three bracketed exposure images on a tripod used to generate ground-truth, and afterwards two additional images are captured while the subject is asked to move, obtaining therefore a input LDR triplet with inter-frame motion and a reference HDR ground-truth image aligned to the central frame. Such capturing approaches are normally limited to small datasets, as this type of capturing is time consuming, and additionally it constrains the motions that can be captured while misalignment might still happen if the subject is not completely still.

For this challenge we introduce a newly curated HDR dataset. This dataset is composed of approximately 1500 training, 60 validation and 201 testing examples. Each example in the dataset is in turn composed of three input LDR images, \ie\,short, medium and long exposures, and a related ground-truth HDR image aligned with the central medium frame. The images are collected from the work of Froelich \etal~\cite{Froehlich14}, where they capture an extensive set of HDR videos using a professional two-camera rig with a semitransparent mirror for the purpose of HDR display evaluation. The contents of those videos include naturally challenging HDR scenes: \eg\,moving light sources, brightness changes over time, high contrast skin tones, specular highlights and bright, saturated colors. As these images lack the necessary LDR input images, similarly to \cite{kalantari19}, we synthetically generate the respective LDR counterparts by following accurate image formation models that include several noise sources \cite{Hasinoff10}.
%\hspace{-4.3mm}

\textbf{Image Formation Model}: In order to model the HDR to LDR step, we use the following pixel measurement model as described in \cite{Hasinoff10}:
\begin{equation}
I_l=\min\left\{\varPhi t/g+I_{0}+n,\,I_{max}\right\},
\label{eq:image-formation}
\end{equation}
where $I_l$ is an LDR image, $\varPhi$ is the scene brightness, $t$ is the exposure time, $g$ is the sensor gain, $I_0$ is the constant offset current, $n$ is the sensor noise and $I_{max}$ denotes the saturation point. For our data processing, we assume $\varPhi$ to be well approximated by the ground-truth HDR images, and produce different LDR images by modifying the exposure time $t$  parameter of any three consecutive frames.

\textbf{Noise Model}: In order to realistically reproduce the characteristic of common LDR images, we include a zero-mean signal whose variance comes from three independent sources, \ie\,photon noise, read noise and analog-to-digital (ADC) gain and quantization noise (for 8-bit LDR images). For pixels under the saturation level, the variance of $n$ reads:
\begin{equation}
\textrm{Var}(n)=\varPhi t/g^{2}+\sigma_{read}^{2}/g^{2}+\sigma_{ADC}^{2}
\label{eq:noise-var}
\end{equation}
Note that first photon-noise term is signal-dependent (normally represented by a Poisson distribution), while the read-noise term is gain-dependent. 

We show in Figure~\ref{fig:dataset-vis} some examples of the HDR  and the synthetically generated LDR images. 

\textbf{Partitions:} We provide training, testing and validation data splits. With our synthetically processed set, we manually discard images to balance the number of frames per scene and to remove undesirable frames, mostly due to \eg dominant presence of lights, lack of inter-frame motion, excessive presence of noise in the HDR image. This leads to roughly 1750 frames within 29 different scenes. The validation and testing splits are obtained randomly from 4 different scenes (\textit{carousel fireworks 02, fireplace 02, fishing longshot, poker travelling slowmotion}) while the other 25 scenes are used for the training set, ensuring that there is no scene overlap between training and testing/validation. This results on a training set short of $1500$ examples, and a validation and testing set of $60$ and $201$ examples respectively.
%TODO 
% Discuss splits, and add some visualizations

\subsection{Challenge Design and Tracks}
This challenge is organized into two different tracks, both of them sharing the same evaluation metrics and ground-truth data. The results from both tracks are thus directly comparable and can explain the performance differences between single and multi-frame HDR imaging. 

\subsubsection{Track 1: Single Frame}
This track evaluates the HDR reconstruction when only a single LDR frame is available. In contrast to the multi-frame approaches, single image HDR methods have only a single exposure (instead of a bracketed set) which is arguably more challenging when recovering under- and over-exposed regions as no information from neighboring frames at different EVs can be leveraged. Similarly, single image denoising poses further challenges than its multiple frame counterpart as noise sources are of zero mean and less observations are available. On the other side, this track does not have to deal with motion related artifacts, \eg\,ghosting, bleeding edges, which are common in the multiframe set-up.

\subsubsection{Track 2: Multiple Frame}
This track evaluates the HDR reconstruction for three differently exposed LDR images (\ie\,short, medium, long) with diverse motion between the respective frames, including camera motion, non-rigid scene motion with an emphasis on complex moving and changing light sources. The bracketed input frames were separated by steps of $2$ or $3$ EV between them, similarly to other existing datasets~\cite{kalantari17}. In order to enable direct comparison between both tracks, the medium LDR frame in Track 2 corresponds to the single-frame LDR input on Track 1 and thus both tracks share the same ground-truth data.

\subsection{Evaluation}
The evaluation of the challenge submissions is based on the computation of objective metrics between the estimated and the ground-truth HDR images. We use the well-known standard peak signal-to-noise ratio (PSNR) both directly on the estimated HDR image and after applying the $\mu$-law tonemapper, which is a simple and canonical operator used widely for benchmarking in the HDR literature \cite{kalantari17,prabhakar20,yan19}. From within these two metrics, we selected PSNR-$\mu$ as the main metric to rank methods in the challenge.

For the PSNR directly computed on the estimated HDR images we normalize the values to be in the range $[0,1]$ using the peak value of the ground-truth HDR image. 

For the PSNR-$\mu$, we apply the following tone-mapping operation $\mathrm{T}(H)$:
\begin{equation}
\label{eq:mu-law}
\mathrm{T}(H)  = \frac{\log(1 + \mu H)}{\log(1 + \mu)}
\end{equation}
where $H$ is the HDR image, and $\mu$ is a parameter that controls the compression, which we fix to $\mu = 5000$ following common HDR evaluation practises. In order to avoid excessive compression due to peak value normalization, for the PSNR-$\mu$ computation we normalize using the $99$ percentile of the ground-truth image followed by a $\tanh$ function to maintain the $\left[0,1\right]$ range.

%-------------------------------------------------------------------------
\section{Results}
\begin{figure}
\centering
\includegraphics[width=1.0\columnwidth]{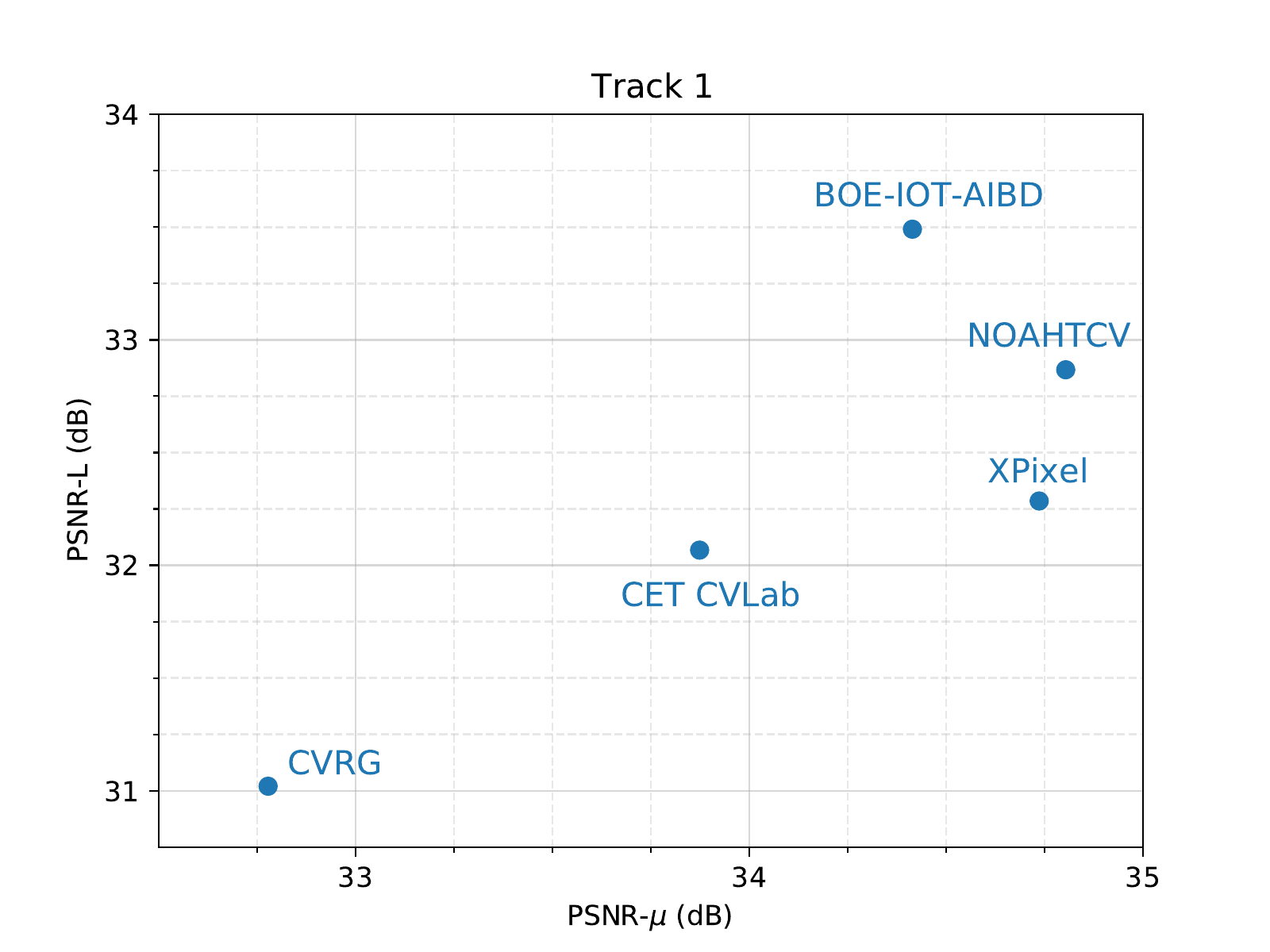}
% Let's start using labels and citing them in the main document
\caption{Combined PSNR-$\mu$ and PSNR values of methods from the \textit{Track 1 (single frame)}.}
\label{fig:track1} 
\includegraphics[width=1.0\columnwidth]{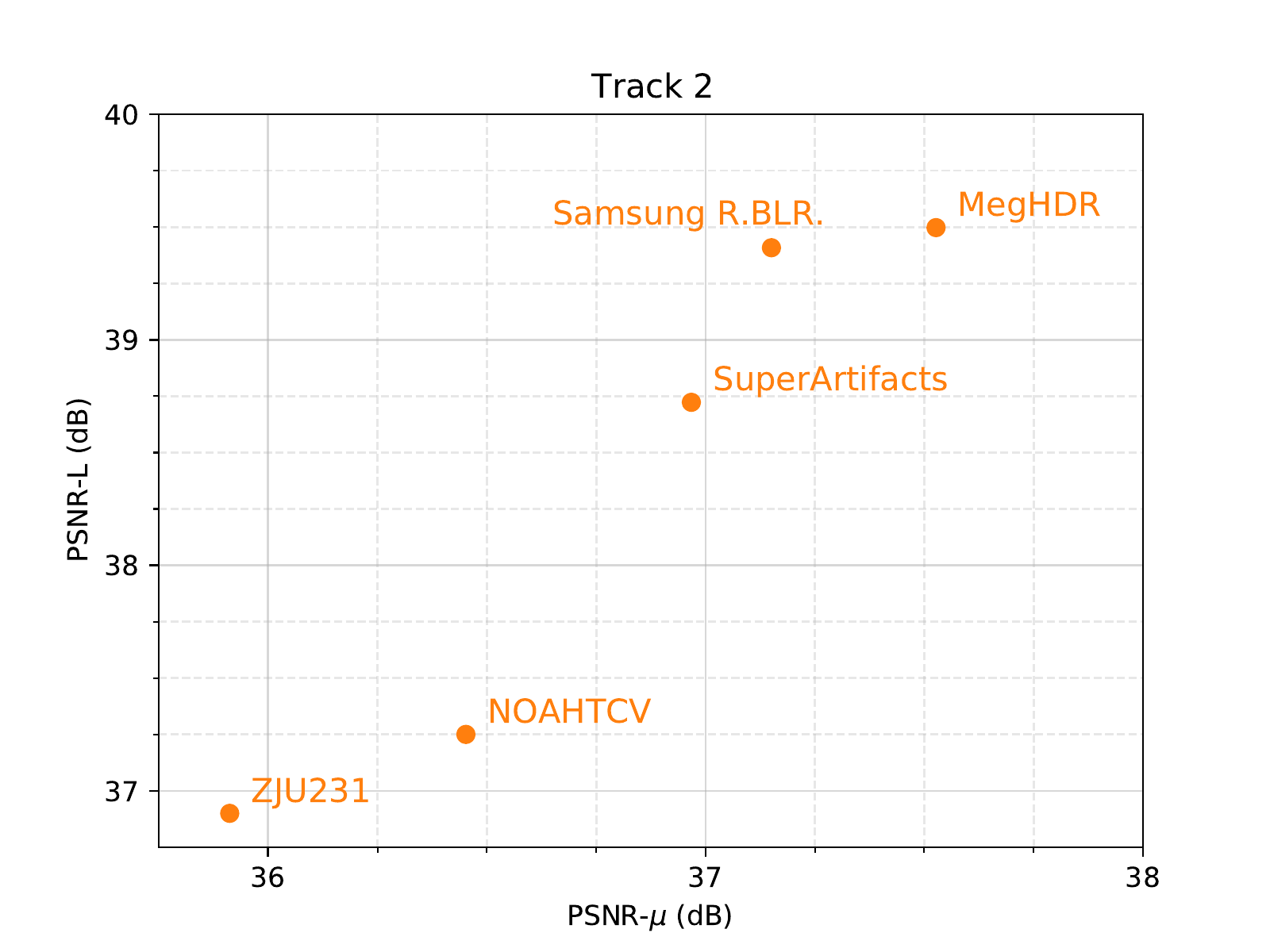}
% Let's start using labels and citing them in the main document
\caption{Combined PSNR-$\mu$ and PSNR values of methods from the \textit{Track 2 (multiple frames)}.}
\label{fig:track2} 

\includegraphics[width=1.0\columnwidth]{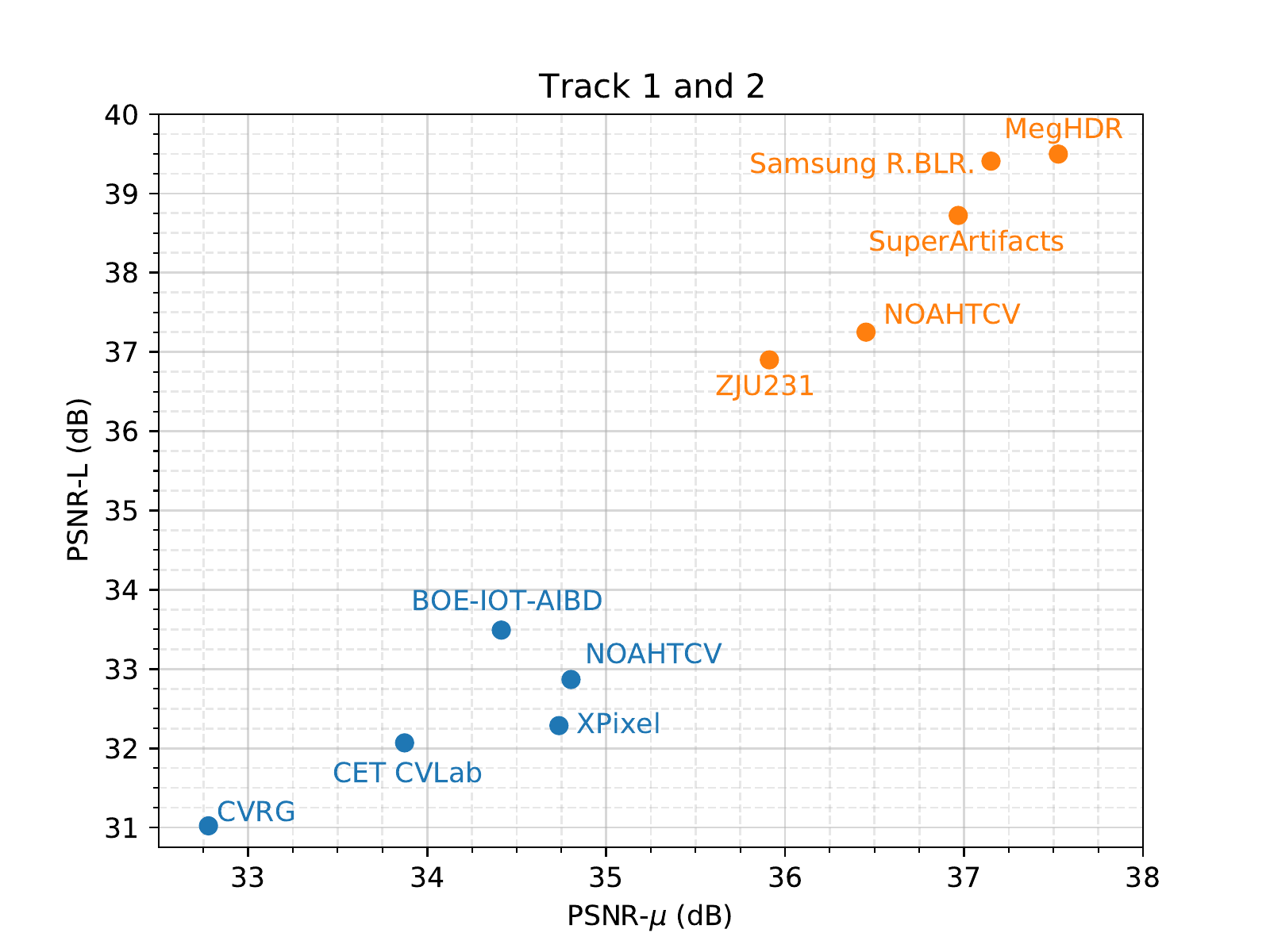}
% Let's start using labels and citing them in the main document
\caption{Combined PSNR-$\mu$ and PSNR values of methods from the \textit{Track 1} (in blue) and \textit{Track 2} (in orange).}
\label{fig:track1-and-2} 
\end{figure}

From  120 registered participants in Track 1, 16 teams participated during the development phase and finally 7 teams entered the final testing phase and submitted results and fact sheets. As for Track 2, from 126 registered participants, 28 teams participated during the development phase and finally 6 teams entered the final testing phase and submitted results and fact sheets. We report the final test phase results in Table~\ref{table:track1-test} and \ref{table:track2-test} for track 1 and 2 respectively. A visualization of both metrics for each track separately can be found in Figure~\ref{fig:track1} and \ref{fig:track2}, and all the results from both tracks are aggregated in Figure~\ref{fig:track1-and-2}.
% , and for completeness we also report the development phase results (\ie\,results for the validation set) in Table REF and REF as well
The methods and the teams that entered the final phase are described in Section~\ref{sec:team-and-methods}, more detailed information about each team and their member's affiliation can be found in Appendix~\ref{ap:teams-and-affiliations}.

\subsection{Main ideas}
In the single frame track, the majority of the proposed architectures consist of several sub-networks which aim to reverse single aspects of the HDR to LDR image pipeline, perhaps inspired by the success of Liu \etal~\cite{liu20}. Variants of the Residual Dense Block \cite{zhang18} are the most commonly used backbone although U-Net style architectures are used by a significant minority. In addition to the standard $\ell_{1}$ loss, some methods also use perceptual colour losses.
 
In the multiple frames track, a major number of solutions are inspired by Yan \etal AHDRNet~\cite{yan19}, with most submissions using their attention mechanism. Most methods also adopt the Dilated Residual Dense Block, although similarly to Track 1, U-Net style architectures with non-dense residual blocks are also present and achieve competitive performance. Ensemble approaches to improve performance via test time augmentations such as flips/transpose~\cite{timofte2016seven} are common among the participants, leading to increases of up to $0.5$ dB. Some submissions aim to explicitly align input images instead of just rejecting unaligned regions with attention, including the first-ranked submission which aligns images using deformable convolutions~\cite{Dai17,liu21_ntire}.

\subsection{Top results}
\textbf{Track 1}: The top two methods (NOAHTCV and XPixel) obtain similar PSNR-$\mu$ scores, only about $0.07$ dB apart, while in terms of PSNR the difference is more noticeable, in the range of $0.6$ dB. BOE-IOT-AIBD comes third in terms of PSNR-$\mu$, at around $0.4$ dB gap to the first position, however they are ranked first in terms of PSNR by a noticeable margin ($0.6$ dB) to the second best-performer in that metric (NOAHTCV). The rest of competing teams obtain scores within $2$ and $1$ dB gaps when compared to the best-performer in terms of PSNR and PSNR-$\mu$ respectively.

\textbf{Track 2}: In this track, both metrics behave similarly and exactly the same ranking is obtained with either of them. The MegHDR team obtains the first position, with a lead of $0.56$ dB in terms of PSNR-$\mu$ and a broader difference of $0.78$ in terms of PSNR when compared against the runner-up team in the leaderboard (SuperArtifacts). NOAHTCV follows at roughly $0.5$ dB and $0.8$ dB performance gap with respect to MegHDR in terms of PSNR-$\mu$ and PSNR respectively. The rest of competing teams obtain scores within $1.6$ and $2.6$ dB gaps when compared to the best-performer in terms of PSNR and PSNR-$\mu$ respectively.

\begin{table*}
\centering
\renewcommand{\arraystretch}{1.5}
\begin{tabular}{l c c c c c c} \toprule
    Team & Username & PSNR-$\mu$ & PSNR & Runtime (s) & GPU & Ensemble  \\ \midrule
    NOAHTCV & noahtcv & $34.804~_{(1)}$ & $32.867~_{(2)}$ & $61.52~_{(5)}$ & Tesla P100 & flips, transpose \\ 
    XPixel & Xy\textunderscore Chen  & $34.736~_{(2)}$ & $32.285~_{(3)}$ &  $0.53~_{(2)}$ & RTX 2080 Ti & - \\
    BOE-IOT-AIBD & chenguannan1981 & $34.414~_{(3)}$ & $33.490~_{(1)}$ & $5.00~_{(4)}$ & Tesla V100 & - \\
    CET CVLab & akhilkashok & $33.874~_{(4)}$ &  $32.068~_{(4)}$ & $0.20~_{(1)}$ & Tesla P100 & flips, rotation \\
    CVRG & sharif\textunderscore apu & $32.778~_{(5)}$ & $31.021~_{(5)}$ & $1.10~_{(3)}$ & GTX 1060 & - \\
    \textit{no processing} & - & $25.266~_{(6)}$ & $27.408~_{(6)}$ & - & - & - \\ 
    \bottomrule
\end{tabular}

\vspace{0.75ex}
\caption{Results and rankings of methods submitted to the \textit{Track 1: Single frame HDR}. Please note that running times are self-reported.}
\label{table:track1-test}
\end{table*}

\begin{table*}
\centering
\renewcommand{\arraystretch}{1.5}
\begin{tabular}{l c c c c c c} \toprule
    Team & Username & PSNR-$\mu$ & PSNR & Runtime (s) & GPU & Ensemble  \\ \midrule
    MegHDR & liuzhen & $37.527~_{(1)}$ & $39.497~_{(1)}$ & $1.35~_{(2)}$ &  RTX 2080 Ti & flips, transpose \\ 
    SuperArtifacts & evelynchee & $36.968~_{(2)}$ & $38.723~_{(2)}$ & $3.80~_{(4)}$ & RTX 2080 Ti & - \\
    NOAHTCV & noahtcv & $36.452~_{(3)}$ &  $37.250~_{(3)}$ & $1.26~_{(1)}$ & Tesla V100 & - \\
    ZJU231 & ZJU231 & $35.912~_{(4)}$ & $36.900~_{(4)}$ & $2.96~_{(3)}$ & RTX 2080 Ti &
    {\renewcommand{\arraystretch}{0.8}
    \begin{tabular}{c}
    flips, rotation,\\
    $\times 4$ models
    \end{tabular}}\\
    {\renewcommand{\arraystretch}{0.8}
    \hspace{-1.8ex}\begin{tabular}{l}
     Samsung Research\\Bangalore$^*$
     \end{tabular}} & AnointedKnight  & \hspace{-3.5ex} $37.151$ & \hspace{-3.2ex}$39.408$ &  \hspace{-6.5mm}$15.77$ & Tesla P40 & flips, transpose \\
    \textit{no processing} & - & $25.266~_{(5)}$ & $27.408~_{(5)}$ & - & - & - \\ 
    \bottomrule
\end{tabular}
\vspace{0.75ex}
\caption{Results and rankings of methods submitted to the \textit{Track 2: Multiple frames HDR}. Please note that running times are self-reported.}
\label{table:track2-test}
\end{table*}

%-------------------------------------------------------------------------
\section{Team and Methods\label{sec:team-and-methods}}
\subsection{NOAHTCV}
NOAHTCV have proposed two methods, one for single frame and one for multi-frame. Both methods are discussed here. 

\begin{figure}[h]
\centering
\includegraphics[width=1.0\columnwidth]{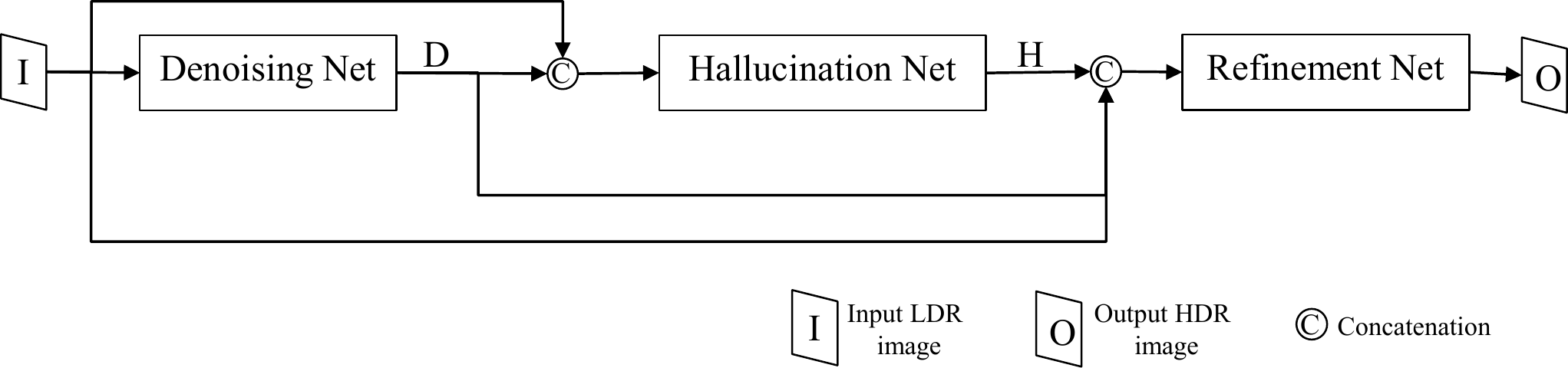}
% Let's start using labels and citing them in the main document
\caption{Architecture of Single Image HDR Reconstruction in a Multi-stage Manner, proposed by the NOAHTCV team.}
\label{fig:noah-single} 
\end{figure}

\textbf{Single Image HDR Reconstruction in a Multi-stage Manner} The team propose a multi-stage method which decomposes the problem into two sub-tasks; denoising and hallucination. The input image, $I$ is first passed through a denoising network to get the denoised image $D$. Both $I$ and $D$ are processed by the hallucination network to obtain $H$. Finally $I$, $D$ and $H$ are fused by a refinement network. The general architecture can be seen in Figure~\ref{fig:noah-single}. 
MIRNet~\cite{zamir20} is employed as the denoising network, while the hallucination network uses masked features as in~\cite{santos20} to reconstruct details in the over-exposed regions. The refinement network is a U-Net equipped with coordinate attention~\cite{hou21}.

\begin{figure}[h]
\centering
\includegraphics[width=1.0\columnwidth]{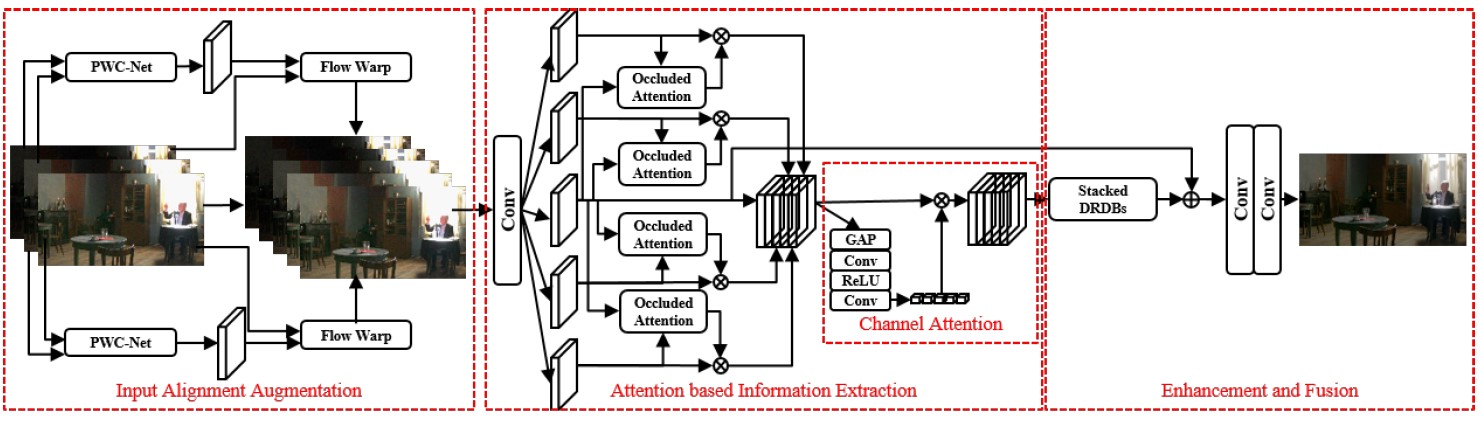}
% Let's start using labels and citing them in the main document
\caption{Architecture of Alignment Augmentation and Multi-Attention Guided HDR Fusion, proposed by the NOAHTCV team.}
\label{fig:noah-multiple} 
\end{figure}

\textbf{Alignment Augmentation and Multi-Attention Guided HDR Fusion} The team propose a three stage method consisting of an Alignment and Augmentation module, an Attention Based Information Extraction module, and an Enhancement and Fusion module. The architecture can be seen in Figure~\ref{fig:noah-multiple}. The Alignment and Augmentation module uses a pretrained PWC-Net~\cite{sun18} to warp the short and long input images with a predicted optical flow. Both the original images and warped images are fed into the network. The Attention Based Information Extraction module employs the occluded attention mechanism from AHDRNet~\cite{yan19} to reduce misalignment distortion. Channel attention is also used on shallow features extracted by a shared convolutional layer to re-weight features generated by different frames. The Enhancement and Fusion module employs the network architecture from AHDRNet~\cite{yan19} with the final sigmoid layer removed.

\subsection{MegHDR}
\begin{figure}[h]
\centering
\includegraphics[width=1.0\columnwidth]{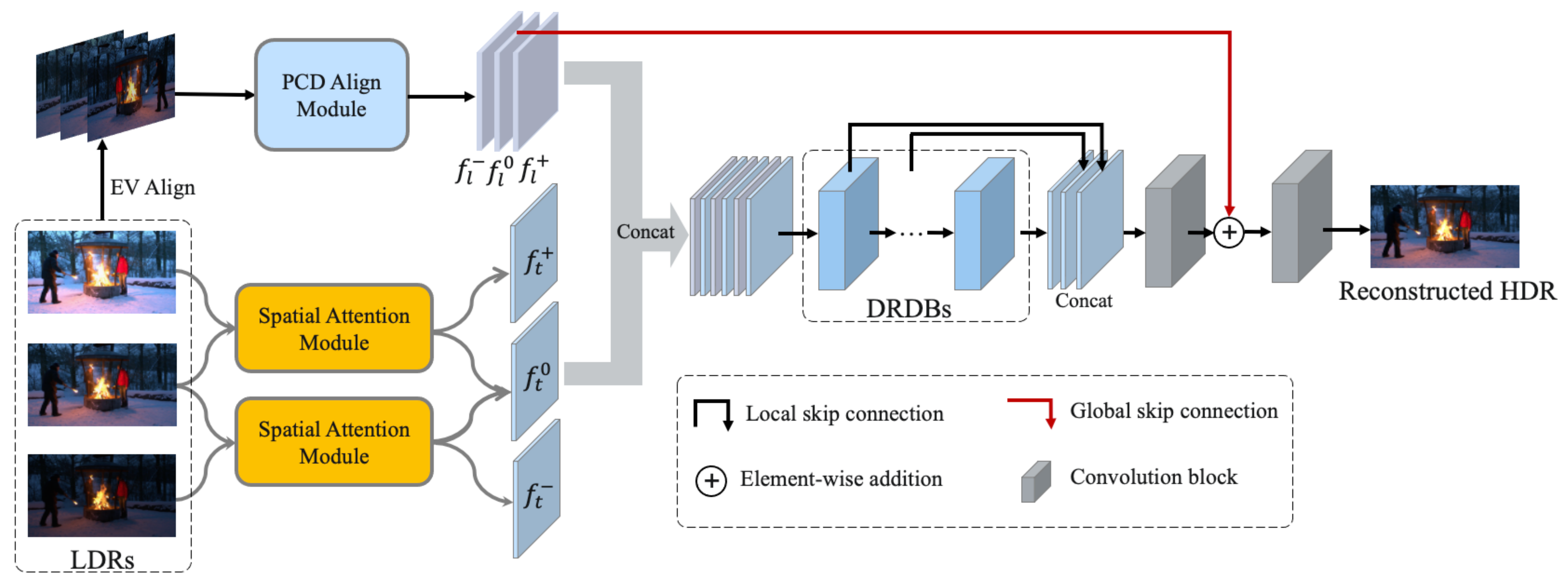}
% Let's start using labels and citing them in the main document
\caption{Architecture of ADNet, proposed by the MegHDR team.}
\label{fig:meghdr} 
\end{figure}

\textbf{ADNet: Attention-guided Deformable Convolutional Networks for High Dynamic Range Imaging} The team propose ADNet~\cite{liu21_ntire}, a novel multi-frame imaging pipeline where the LDR images and their corresponding gamma-corrected images are processed separately, instead of being concatenated together. This is motivated by the intuition that images in the LDR domain are helpful for detecting noisy or saturated regions, while images in the HDR domain help to detect misalignment. The PCD align module aligns the gamma corrected images using pyramid, cascading and deformable convolutions based on EDVR~\cite{wang19}.  The spatial attention module suppresses undesired saturation and noisy regions in the LDR images while highlighting the regions useful for fusion. The resulting features are concatenated and processed by dilated residual dense blocks (DRBDs) as in AHDRNet~\cite{yan19}. The architecture can be seen in Figure~\ref{fig:meghdr}. 

\subsection{XPixel}
\begin{figure}[h]
\centering
\includegraphics[width=1.0\columnwidth]{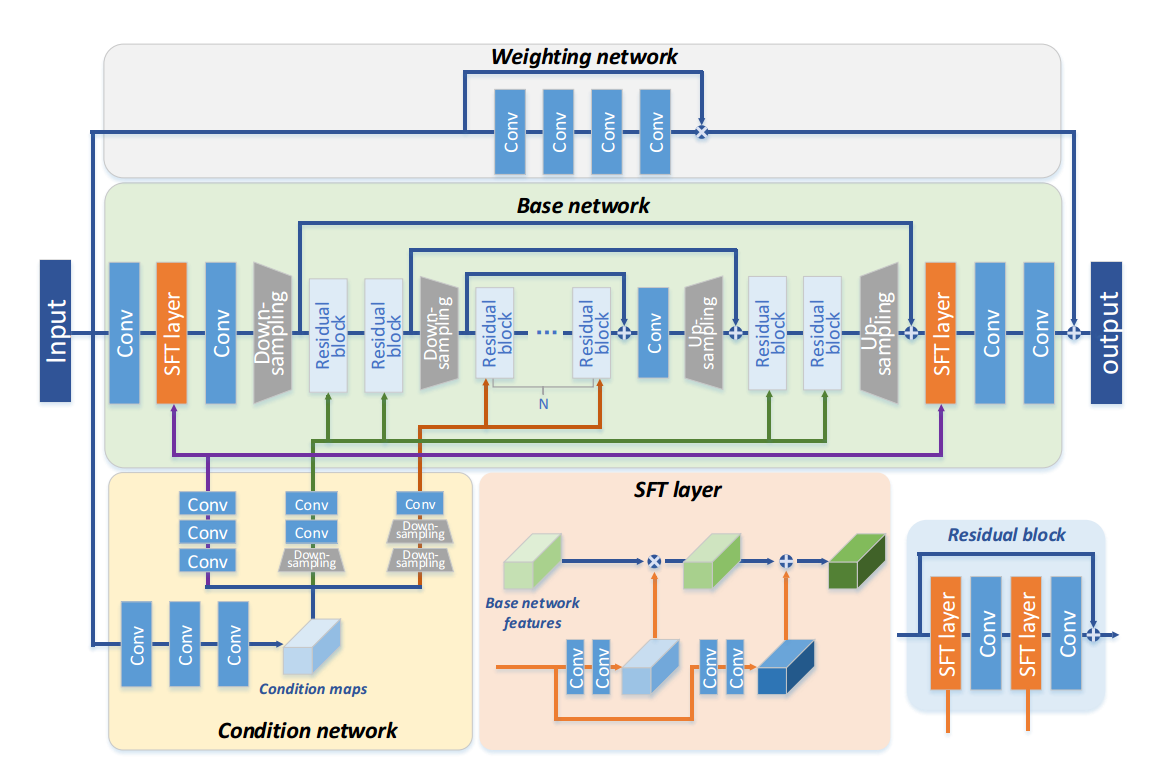}
% Let's start using labels and citing them in the main document
\caption{Architecture of HDRUNet: Single Image HDR Reconstruction with Denoising and Dequantization, proposed by the XPixel team.}
\label{fig:xpixel} 
\end{figure}

\textbf{HDRUNet: Single Image HDR Reconstruction with Denoising and Dequantization} The team propose HDRUNet \cite{chen2021hdrunet}, which consists of three sub-networks: the base network, the condition network and the weighting network. The architecture can be seen in Figure \ref{fig:xpixel}. The base network is a U-Net style encoder-decoder model. The condition network and spatial feature transform (SFT) layers~\cite{wang18} are introduced to achieve adaptive modulation based on the features being processed. Besides, inspired by~\cite{Eilersten17}, a mask is calculated for the global residual, as adding it directly is sub-optimal. Finally, a $\tanh\: \ell_1$ loss function is adopted to balance the impact of over-exposed values and well-exposed values on the network learning.
{\let\thefootnote\relax\footnotetext{%
\hspace{-5mm}$^*$ Incomplete submission (no reproducibility) thus not in the challenge ranking.}}

\subsection{BOE-IOT-AIBD}
\begin{figure}[h]
\centering
\includegraphics[width=1.0\columnwidth]{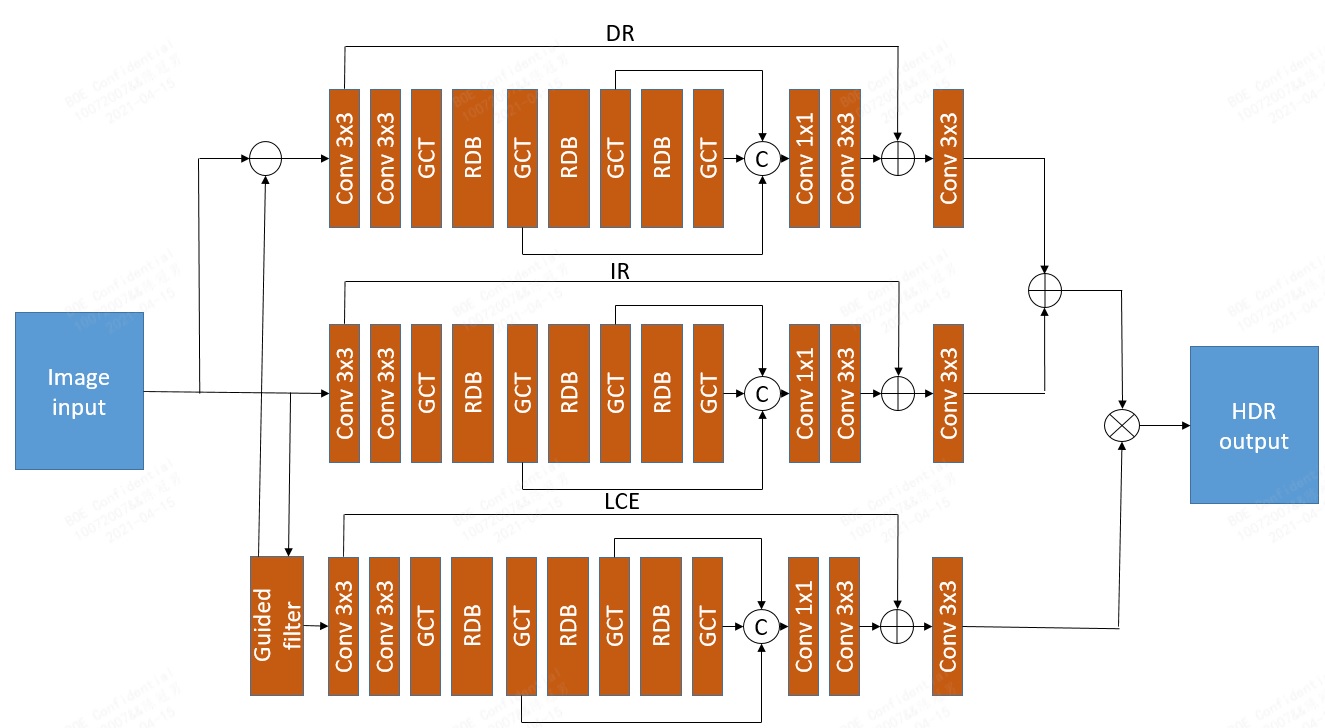}
% Let's start using labels and citing them in the main document
\caption{Architecture of Task-specific Network based on Channel Adaptive RDN, proposed by the BOE-IOT-AIBD team.}
\label{fig:boe} 
\end{figure}

\textbf{Task-specific Network based on Channel Adaptive RDN} The team propose a method~\cite{chen21_ntire_RDN} which consists of three sub-networks which each perform a different task: Image Reconstruction (IR), Detail Restoration (DR) and Local Contrast Enhancement (LCE)~\cite{Kim2020JSI}. The IR network reconstructs the coarse HDR image from the input LDR image. The DR network can further refine the image details by adding its output to the coarse HDR output of IR. Finally the LCE network predicts a luminance equalization mask which is multiplied by the refined HDR image for contrast adjustment. The total architecture can be seen in Figure~\ref{fig:boe}. All three sub-networks use the same backbone, named the Channel Adaptive RDN. This consists of the standard Residual Dense Network~\cite{zhang18} with the Gate Channel Transformation layer~\cite{yang20} added to each RDB block.

\subsection{SuperArtifacts}
\begin{figure}[h]
\centering
\includegraphics[width=1.0\columnwidth]{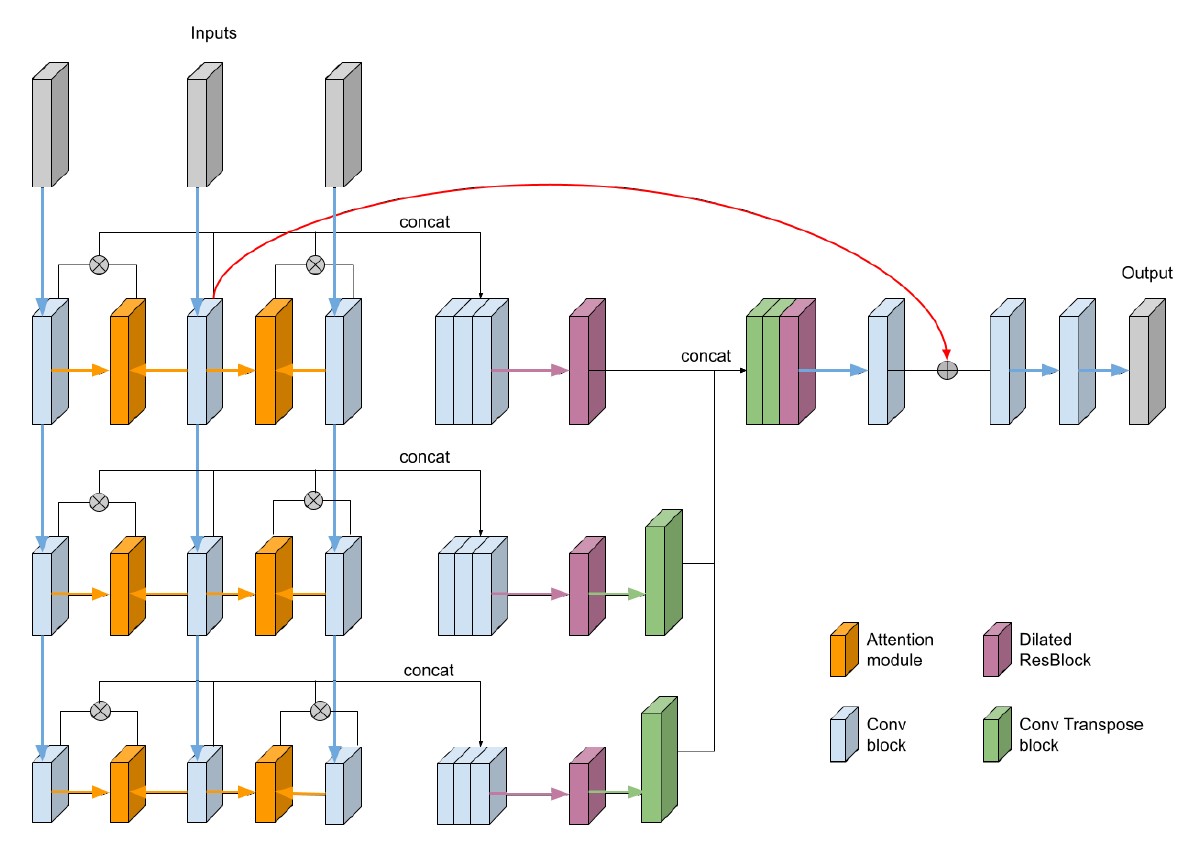}
% Let's start using labels and citing them in the main document
\caption{Architecture of Multi-Level Attention on Multi-Exposure Frames for HDR Reconstruction, proposed by the SuperArtifacts team.}
\label{fig:super} 
\end{figure}

\textbf{Multi-Level Attention on Multi-Exposure Frames for HDR Reconstruction} The team propose a multi-level architecture which processes and merges features at three different resolutions. On top of the architecture of  AHDRNet~\cite{yan19}, the model encodes the frames into three levels, with each feature being half the resolution of the previous level. This increases the receptive field and helps to better handle large foreground motion. At each level, the attention mechanism is used to identify which regions to use from the long and short exposure frames. The features at each level are merged independently first before being upsampled back to the original resolution. The features from all three levels are then merged together using some fusion blocks to generate the final HDR image. The architecture can be seen in Figure~\ref{fig:super}.

\subsection{CET-CVLAB}
\begin{figure}
\centering
\includegraphics[width=1.0\columnwidth]{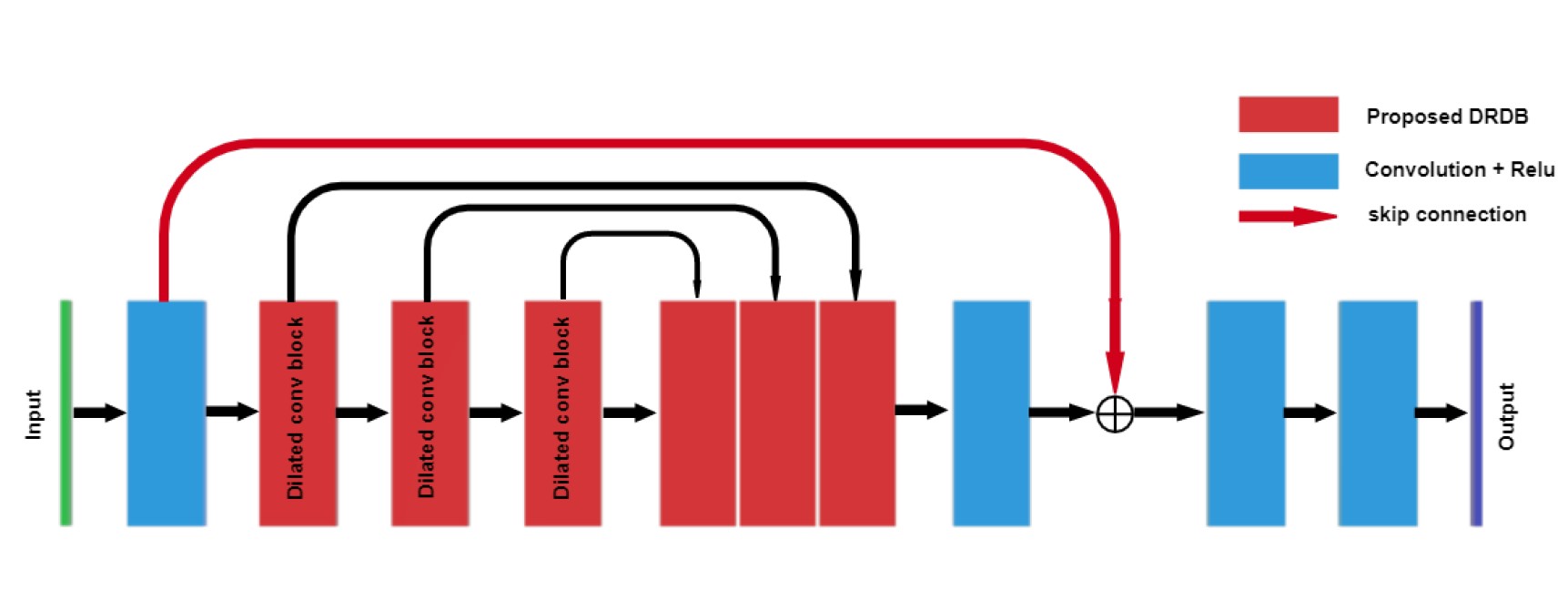}
% Let's start using labels and citing them in the main document
\caption{Architecture of Single Image HDR Synthesis with Densely Connected Dilated ConvNet, proposed by the CET-CVLAB team.}
\label{fig:cet} 
\end{figure}

\textbf{Single Image HDR Synthesis with Densely Connected Dilated ConvNet} The team propose an architecture which consists of a densely connected stack of dilated residual dense blocks (DRDBs)~\cite{akhil21_ntire}. The dilation rate of convolutional layers used within the proposed DRDB progressively grows from 1 to 3 and then progressively decreases from 3 to 1. The DRDBs themselves are also connected as shown in Figure~\ref{fig:cet} to improve the representation capability of the network.

\subsection{CVRG}
\begin{figure}[h]
\centering
\includegraphics[width=1.0\columnwidth]{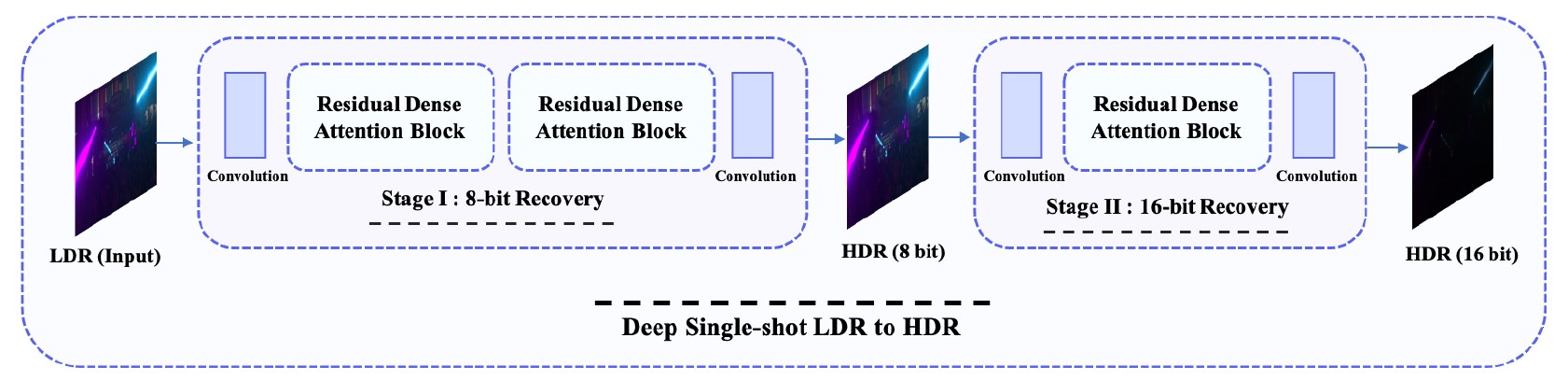}
% Let's start using labels and citing them in the main document
\caption{Architecture of Deep Single-Shot LDR to HDR, proposed by the CVRG team.}
\label{fig:cvrg} 
\end{figure}

\begin{figure}[h]
\centering
\includegraphics[width=1.0\columnwidth]{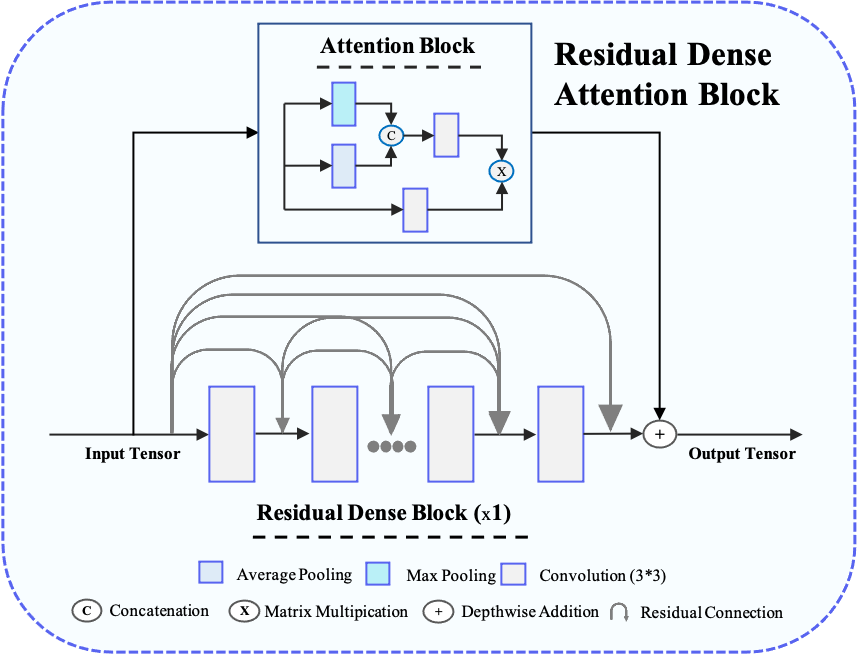}
% Let's start using labels and citing them in the main document
\caption{Residual Dense Attention Block, proposed by the CVRG team.}
\label{fig:rdab} 
\end{figure}

\textbf{Deep Single-Shot LDR to HDR} The team propose a two stage method~\cite{sharif21_ntire}: Stage I (inspired by~\cite{sharif20}) performs denoising and recovers the 8-bit HDR image from the single LDR input; Stage II tonemaps the image into the linear domain and recovers the 16-bit HDR image. The architecture can be seen in Figure~\ref{fig:cvrg}. The team proposes the Residual Dense Attention Block (RDAB) as the building block of the model. The RDAB, which combines the residual dense block and the spatial attention module, can be seen in Figure~\ref{fig:rdab}.

\subsection{ZJU231}
\begin{figure}[h]
\centering
\includegraphics[width=1.0\columnwidth]{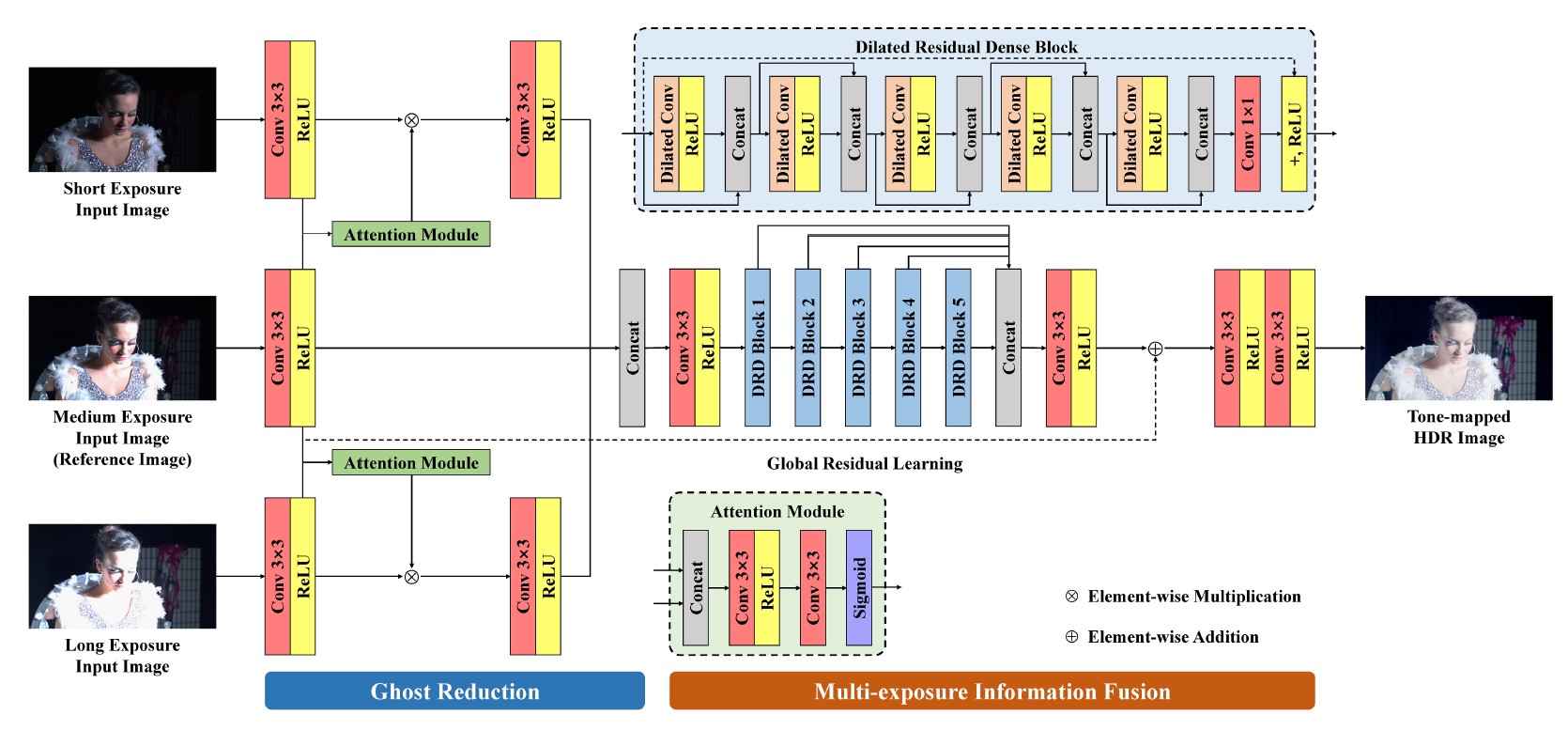}
% Let's start using labels and citing them in the main document
\caption{Architecture of Reference-Guided Multi-Exposure Fusion Network for HDR Imaging, proposed by the ZJU231 team.}
\label{fig:zju} 
\end{figure}

\textbf{Reference-Guided Multi-Exposure Fusion Network for HDR Imaging} The team propose a two-stage architecture which consists of the ghost reduction sub-network and the multi-exposure information fusion sub-network. Inspired by AHDRNet~\cite{yan19}, the ghost reduction sub-network uses the reference image to generate an attention map for the short and long exposure images. The extracted features are guided via element-wise multiplication with the attention maps. The guided features are concatenated and merged by the fusion sub-network, which consists of five DRDBs followed by three convolutions, as shown in Figure~\ref{fig:zju}.  

\subsection{Samsung Research Bangalore}
\begin{figure}[h]
\centering
\includegraphics[width=1.0\columnwidth]{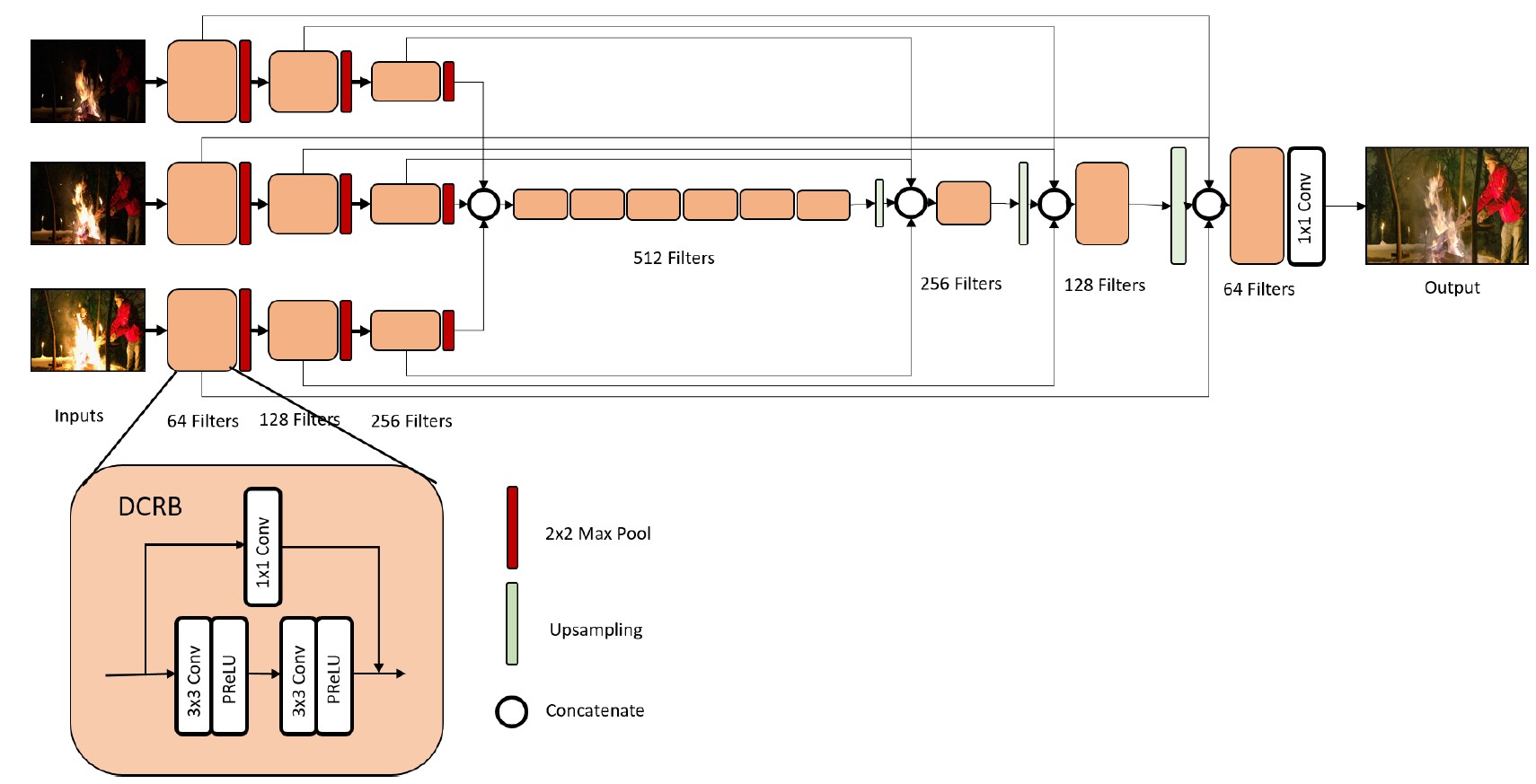}
% Let's start using labels and citing them in the main document
\caption{Architecture of HDR Merging using Multi Branch Residual Networks, proposed by the Samsung Research Bangalore team}
\label{fig:samsung} 
\end{figure}

\textbf{HDR Merging using Multi Branch Residual Networks} The team propose a multi-branch U-Net architecture inspired by~\cite{wu18} and~\cite{green19} which consists of an encoder, a residual body and a decoder as seen in Figure~\ref{fig:samsung}. The building blocks of the network are Double Convolutional Residual Blocks (DCRB). This consists of two convolutions with prelu activations and the input is skipped to the output using a 1x1 convolution. 

Each input image is processed with separate branches. The encoder consists of three blocks which successively downsample the input image. The features are then concatenated and processed using six residual blocks, followed by three decoder blocks which upsample the image back to full resolution. There are skip connections between all three of the encoder and decoder blocks. Self-ensembling strategy by averaging 8 ensembles created using flip and transpose operations are used to further improve the results.

\section*{Acknowledgements}

We thank the NTIRE 2021 sponsors: Huawei,
Facebook Reality Labs, Wright Brothers Institute, MediaTek, OPPO and ETH Zurich (Computer Vision Lab).

% \section*{A. Teams and Affiliations}
\appendix\section{Teams and Affiliations}
\label{ap:teams-and-affiliations}
% First ranked
\subsection*{NOAHTCV}
\noindent \textbf{Title:} Single Image HDR Reconstruction in a Multi-stage Manner / Alignment Augmentation and Multi-Attention Guided HDR Fusion
\newline
\textbf{Members:} Xian Wang$^{1}$ (\href{wangxian10@huawei.com}{wangxian10@huawei.com}), Yong Li$^{1}$, Tao Wang$^{1}$ and Fenglong Song$^{1}$
\newline
\textbf{Affiliations:} $^{1}$Huawei Noah's Ark Lab

\subsection*{MegHDR}
\noindent \textbf{Title:} ADNet: Attention-guided Deformable Convolutional Networks for High Dynamic Range Imaging \newline
\textbf{Members:} Zhen Liu$^{1}$ (\href{mailto:liuzhen03@megvii.com}{liuzhen03@megvii.com}) Wenjie Lin$^{1}$, Xinpeng Li$^{1}$, Qing Rao$^{1}$, Ting Jiang$^{1}$, Mingyan Han$^{1}$, Haoqiang
Fan$^{1}$, Jian Sun$^{1}$ and Shuaicheng Liu$^{1}$
\newline
\textbf{Affiliations:} $^{1}$Megvii Technology

% 2nd ranked
\subsection*{XPixel}
\noindent \textbf{Title:} HDRUNet: Single Image HDR Reconstruction with Denoising and Dequantization
\newline
\textbf{Members:} Xiangyu Chen$^{1}$ (\href{mailto:chxy95@gmail.com}{chxy95@gmail.com}), Yihao Liu$^{1}$, Zhengwen Zhang$^{1}$, Yu Qiao$^{1}$, Chao Dong$^{1}$ 
\newline
\textbf{Affiliations:} $^{1}$Shenzhen Institutes of Advanced Technology, Chinese Academy
of Sciences

\subsection*{SuperArtifacts}
\noindent \textbf{Title:} Multi-Level Attention on Multi-Exposure Frames for HDR Reconstruction
\newline
\textbf{Members:} Evelyn Yi Lyn Chee$^{1}$  (\href{mailto:evelyn.chee@bst.ai}{evelyn.chee@bst.ai}),  Shanlan Shen$^{1}$ , Yubo Duan$^{1}$ 
\newline
\textbf{Affiliations:} $^{1}$Black Sesame Technologies (Singapore)

% 3rd ranked
\subsection*{BOE-IOT-AIBD}
\noindent \textbf{Title:} Task-specific Network based on Channel Adaptive RDN
\newline
\textbf{Members:} Guannan Chen$^{1}$  (\href{mailto:chenguannan@boe.com.cn}{chenguannan@boe.com.cn}), Mengdi Sun$^{1}$, Yan Gao$^{1}$, Lijie Zhang$^{1}$
\newline
\textbf{Affiliations:} $^{1}$BOE Technology Co., Ltd.
% NOAHTCV already introduced

% 4th ranked
\subsection*{CET-CVLAB}
\noindent \textbf{Title:} Single Image HDR Synthesis with Densely Connected Dilated ConvNet
\newline
\textbf{Members:} Akhil K A$^{1}$ (\href{akhil.93.ka@gmail.com}{akhil.93.ka@gmail.com}), Jiji C V$^{1}$
\newline
\textbf{Affiliations:} $^{1}$College of Engineering Trivandrum, India

\subsection*{ZJU231}
\noindent \textbf{Title:} Reference-Guided Multi-Exposure Fusion Network for HDR Imaging \newline
\textbf{Members:} Chenjie Xia$^{1,2}$ (\href{mailto:chenjiexia@zju.edu.cn}{chenjiexia@zju.edu.cn}), Bowen Zhao$^{1,2}$ (\href{mailto:bowenzhao@zju.edu.cn}{bowenzhao@zju.edu.cn}), Zhangyu Ye (\href{mailto:qiushizai@zju.edu.cn}{qiushizai@zju.edu.cn}), Xiwen Lu, Yanpeng Cao$^{1,2}$, Jiangxin Yang$^{1,2}$,
Yanlong Cao$^{1,2}$
\newline
\textbf{Affiliations:} $^{1}$State Key Laboratory of Fluid Power and Mechatronic Systems, Zhejiang University, $^{2}$Key Laboratory of Advanced Manufacturing Technology, Zhejiang University

% 5th ranked
\subsection*{CVRG}
\noindent \textbf{Title:} Deep Single-Shot LDR to HDR
\newline
\textbf{Members:} S M A Sharif$^{2}$ (\href{mailto:sma.sharif.cse@ulab.edu.bd}{sma.sharif.cse@ulab.edu.bd}), Rizwan Ali Naqvi$^{1}$, Mithun Biswas, and Sungjun Kim
\newline
\textbf{Affiliations:} $^{1}$Sejong University, South Korea, $^{2}$Rigel-IT, Bangladesh

% Outside of
\subsection*{Samsung Research Bangalore}
\noindent \textbf{Title:}HDR Merging using Multi Branch Residual Networks
\newline
\textbf{Members:} Green Rosh K S$^{1}$  (\href{mailto:greenrosh.ks@samsung.com}{greenrosh.ks@samsung.com}), Sachin Deepak Lomte$^{1}$, Nikhil Krishnan$^{1}$ , B H Pawan Prasad$^{1}$ 
\newline
\textbf{Affiliations:} $^{1}$Samsung R\&D Institute India Bangalore (SRI-B)

%-------------------------------------------------------------------------
{\small
\bibliographystyle{ieee_fullname}
\bibliography{egbib}
}

\end{document}